\documentclass[10pt,twocolumn,letterpaper]{article}

\usepackage[pagenumbers]{iccv}
\definecolor{iccvblue}{rgb}{0.21,0.49,0.74}
\usepackage[pagebackref,breaklinks,colorlinks,allcolors=iccvblue]{hyperref}

\usepackage{array}
\usepackage[dvipsnames]{xcolor}
\usepackage{tcolorbox}
\usepackage{makecell}
\usepackage{multirow}
\usepackage[usestackEOL]{stackengine}
\usepackage{colortbl}
\usepackage{pifont}
\usepackage{float}

\title{Motion-Aware Generative Frame Interpolation}

\author{%
  \hspace{-5mm} Guozhen Zhang$^{1,\ddag,*}$ \quad Yuhan Zhu$^{1,\ddag}$ \quad Yutao Cui$^2$ \quad Xiaotong Zhao$^2$ \quad Kai Ma$^2$ \quad Limin Wang$^{1,3,\dag}$ \\
  $^1$State Key Laboratory for Novel Software Technology, Nanjing University\\
  $^2$Platform and Content Group (PCG), Tencent \quad $^3$Shanghai AI Lab\\
  \\
  \newline \textbf{\url{https://mcg-nju.github.io/MoG_Web/}}\\
}
\begin{document}
\twocolumn[{
            \renewcommand\twocolumn[1][]{#1}
            \vspace{-1em}
            \maketitle
            \vspace{-1em}
            \begin{center}
                \vspace{-20pt}
                \centering
                \includegraphics[width=1.0\textwidth]{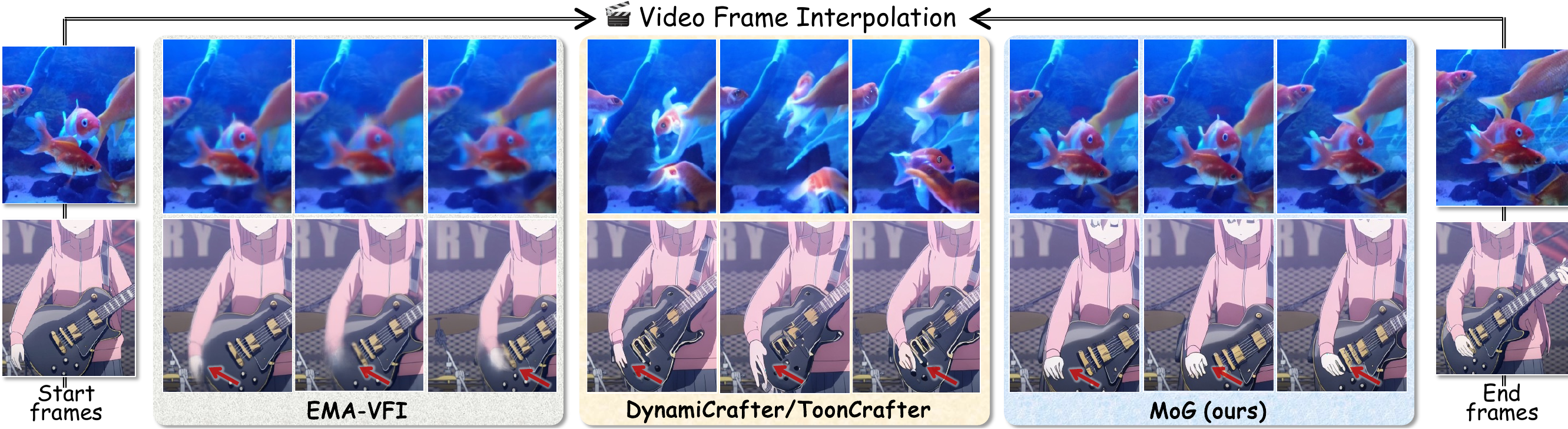}
                \vspace{-7mm}
                \captionof{figure} {
                \textbf{Examples of frame interpolation in real-world and animation scenes.} Compared to other methods, our approach, MoG, exhibits superior stability in motion and consistency in appearance details. 
                % Images are cropped from their original resolution for clearer demonstration.
                % For optimal viewing, please zoom in.
                }\vspace{-1mm}
                \label{fig:teaser}
            \end{center}
        }]

\newcommand\blfootnote[1]{%
\begingroup 
\renewcommand\thefootnote{}\footnote{#1}%
\addtocounter{footnote}{-1}%
\endgroup 
}
{
    \blfootnote{*Work is done during internship at Tencent PCG.~~\ddag Equal contribution.~~\dag Corresponding author  (lmwang@nju.edu.cn).
}
}
        
\begin{abstract}

Flow-based frame interpolation methods ensure motion stability through estimated intermediate flow but often introduce severe artifacts in complex motion regions. 
Recent generative approaches, boosted by large-scale pre-trained video generation models, show promise in handling intricate scenes. 
However, they frequently produce unstable motion and content inconsistencies due to the absence of explicit motion trajectory constraints. 
To address these challenges, we propose \textbf{Mo}tion-aware \textbf{G}enerative frame interpolation (MoG) that synergizes intermediate flow guidance with generative capacities to enhance interpolation fidelity. 
Our key insight is to simultaneously enforce motion smoothness through flow constraints while adaptively correcting flow estimation errors through generative refinement. 
Specifically, we first introduce a dual guidance injection that propagates condition information using intermediate flow at both latent and feature levels, aligning the generated motion with flow-derived motion trajectories.  
Meanwhile, we implemented two critical designs, encoder-only guidance injection and selective parameter fine-tuning, which enable dynamic artifact correction in the complex motion regions. 
Extensive experiments on both real-world and animation benchmarks demonstrate that MoG outperforms state-of-the-art methods in terms of video quality and visual fidelity. 
Our work bridges the gap between flow-based stability and generative flexibility, offering a versatile solution for frame interpolation across diverse scenarios.
\end{abstract}    
\section{Introduction}
\label{sec:intro}

Video Frame Interpolation (VFI), which aims to synthesize intermediate frames between two input frames, has garnered significant attention in recent years due to its capacity for enhancing video frame rates in video post-processing. 
Flow-based VFI methods~\cite{RIFE,EMA-VFI,zhang2024vfimamba,liu2024sparse,kong2022ifrnet} predominantly rely on  estimating the motion between input frames—termed intermediate flow—to warp condition information and generate intermediate frames. 
Conventionally, the generated intermediate frames follow the motion trajectories encoded in the estimated intermediate optical flow, yielding temporally coherent video sequences.
However, in scenarios involving complex motions such as object deformations, accurate intermediate flow estimation becomes infeasible, leading to pronounced artifacts in corresponding regions.
As demonstrated in \cref{fig:teaser} by the flow-based method EMA-VFI~\cite{EMA-VFI}, while the overall results exhibit temporal smoothness, complex regions manifest blurring and ghosting artifacts.

Recent advancements~\cite{tooncrafter,dynamicrafter,GI,TRF} have increasingly focused on leveraging video generation models~\cite{dynamicrafter,SVD} for frame interpolation, capitalizing on their robust generative capacities in dynamic scenes. While notable improvements have been demonstrated in complex scenarios~\cite{tooncrafter}, current approaches rely exclusively on generative models to infer inter-frame correspondences—a capability that remains underdeveloped during generative pre-training. Consequently, these methods often produce unstable motion patterns and inconsistencies with input frames due to the lack of explicit motion trajectory constraints, as illustrated by DynamiCrafter~\cite{dynamicrafter} and ToonCrafter~\cite{tooncrafter} in \cref{fig:teaser}.

 In this work, we introduce a new framework, \textbf{Mo}tion-aware \textbf{G}enerative frame interpolation (MoG), which integrates intermediate flow with generative capacities to to enhance interpolation fidelity. 
 MoG is designed to to bridge the gap between flow-based stability and generative flexibility, offering a versatile solution for frame interpolation across diverse scenarios.
Our core idea is to enforce motion smoothness via flow constraints while rectifying complex motion regions through generative refinement simultaneously.
To attain this objective, we tackle two crucial questions: how to seamlessly incorporate flow constraints into the generative model, and how to endow the generative model with the ability to dynamically correct flow errors.

For the first question, to ensure that the generated motion trajectories are guided by the estimated intermediate flow, we introduce dual guidance injection. 
At both the latent and feature levels, we warp the information of the input frames using the intermediate flow and propagate it into the generation process of intermediate frames, serving as an explicit motion guidance for inferring motion.
Notably, compared with ControlNet-like designs~\cite{zhang2023adding,zhu2024generative}, our design requires no additional parameters and thereby better preserves the pre-trained capabilities. 

To address the second challenge, we introduce two critical designs: encoder-only guidance injection and selective parameter fine-tuning. 
Specifically, guidance is exclusively injected at the encoder stage of the generative model, enabling the decoder to adaptively adjust the generation process. 
Furthermore, we only fine-tune the spatial layers to adapt to guidance fusion while preserving motion modeling capabilities in temporal layers.

To thoroughly evaluate the versatility of our approach, we adapt MoG for both real-world and animation scenes. Experimental results demonstrate that MoG substantially outperforms existing generative interpolation models in both domains. Specifically, the interpolated videos generated by MoG exhibit superior motion stability and improved content consistency, as visually demonstrated in \cref{fig:teaser}. Our contributions are summarized as follows:
\begin{itemize}
    \item We propose a novel frame interpolation framework, MoG, which is the first to bridge the gap between flow-based stability and generative flexibility.
    \item We introduce dual-level guidance injection to constrain the generated motion with the motion trajectories derived from the flow.
    \item We implement encoder-only guidance injection and selective parameter fine-tuning to endow the generative model with the ability to dynamically correct flow errors.
    \item Experimental results showing that MoG significantly outperforms existing generative frame interpolation methods in both qualitative and quantitative aspects.
\end{itemize}

\section{Related Work}

\begin{figure*}[t]
  \centering
  \includegraphics[width=0.8\textwidth]{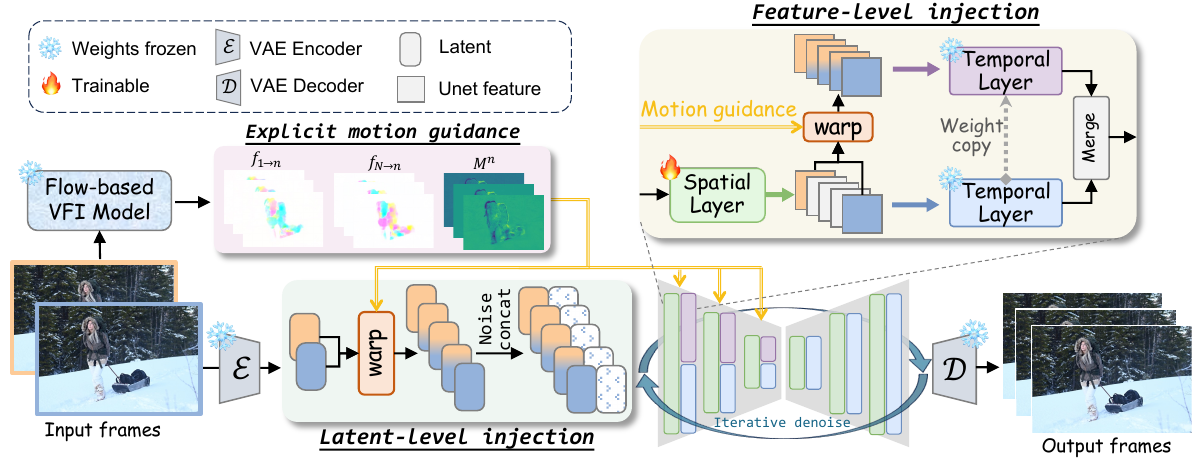}
  \caption{\textbf{Overview of MoG}. MoG consists of two parts. First, it extracts the intermediate flow between input frames. Subsequently, this guidance is seamlessly injected into the generative model at both the latent and feature levels. Meanwhile, the generative model would adaptively rectifying the errors by two crucial designs, namely, encoder-only guidance injection and selective parameter fine-tuning.}
  \label{fig:pipeline}
  \vspace{-3mm}
\end{figure*}

\subsection{Flow-based Frame Interpolation}
Flow-based video interpolation, explicitly estimating the intermediate flow from the input frames to intermediate frames, have become dominant in deterministic frame interpolation~\cite{ming2024survey}. It can be broadly categorized into two classes based on how the intermediate optical flow is derived. The first class~\cite{hu2022many,niklaus2020softmax,niklaus2018context,niklaus2023splatting,bao2019depth,jiang2018super} utilized pre-trained optical flow models to obtain the intermediate flow either directly or through refinement. For instance, SoftSplat~\cite{niklaus2020softmax} linearly adjusted the bidirectional flow estimated by PWC-Net~\cite{sun2018pwc} to represent the intermediate flow and employs an improved forward warping to aggregate information. The second class~\cite{EMA-VFI,RIFE,zhang2024vfimamba,liu2024sparse,kong2022ifrnet,li2023amt,park2023biformer,lu2022videot} modeled the correspondence information of the input frames to directly predict the intermediate flow, offering greater flexibility and task-oriented modeling capacity compared to the first approach. RIFE~\cite{RIFE} demonstrated that simple convolutional layers can effectively predict the intermediate flow, achieving impressive efficiency. Similarly, EMA-VFI~\cite{EMA-VFI} enhanced the flow prediction by explicitly modeling the dynamics between frames through inter-frame cross-attention. However, when confronted with complex motion scenarios, both of classes often exhibit significant blurring and ghosting artifacts. In this work, we leverage the intermediate flow  to enhance the temporal smoothness of generative models. Meanwhile, we utilize generative models to rectify the errors of the intermediate flow in complex scenarios. % We specifically utilize EMA-VFI~\cite{EMA-VFI} to propose intermediate flow due to its high performance and efficiency.

\subsection{Generative Frame Interpolation}
Recent work has begun to explore the use of large-scale pre-trained video generation models~\cite{SVD,dynamicrafter}, which excel at generating videos in complex dynamic scenes, for the VFI task. Current generative frame interpolation methods can be categorized into two types: the first~\cite{TRF,GI,zhu2024generative,yang2024vibidsampler} employed pre-trained generative models to perform image-to-video tasks conditioned on the initial and final frames, subsequently merging the resulting videos to create the final interpolated frame. For example, GI~\cite{GI} enhanced the motion stability by controlling the consistency of temporal correlations across the two generation processes. The second category~\cite{dynamicrafter,tooncrafter,liu2024sparse,voleti2022mcvd,zhou2024storydiffusion,ldmvfi,jain2024video,tanveer2024motionbridge} focused on fine-tuning video generation models specifically for interpolation, by integrating information from input frames into the model’s architecture and optimizing it for end-to-end interpolation. DynamiCrafter~\cite{dynamicrafter} was trained for real-world interpolation, while ToonCrafter~\cite{tooncrafter} was tailored for animated scenes. Although all these methods have demonstrated significant improvements in generating complex scenarios, they do not explicitly consider the correspondence between input frames, which complicates the motion inference of generative models. In contrast, we are the first to explicitly introduce the motion guidance to enhance the motion smoothness of generative models and our method achieves superior video quality and fidelity. Currently, there are also some works~\cite{yang2023rerender,ni2023conditional,liang2024movideo,wang2024levitor} that use optical flow to assist generation in other tasks. They typically require the accurate optical flow between all adjacent frames to aid the generation process, whereas we only use the coarse intermediate flow between the two input frames to smooth the motion. Meanwhile, our method could dynamically correct the errors in the intermediate flow.
\section{Preliminaries}

\begin{figure*}[t]
  \centering
  \includegraphics[width=1.0\textwidth]{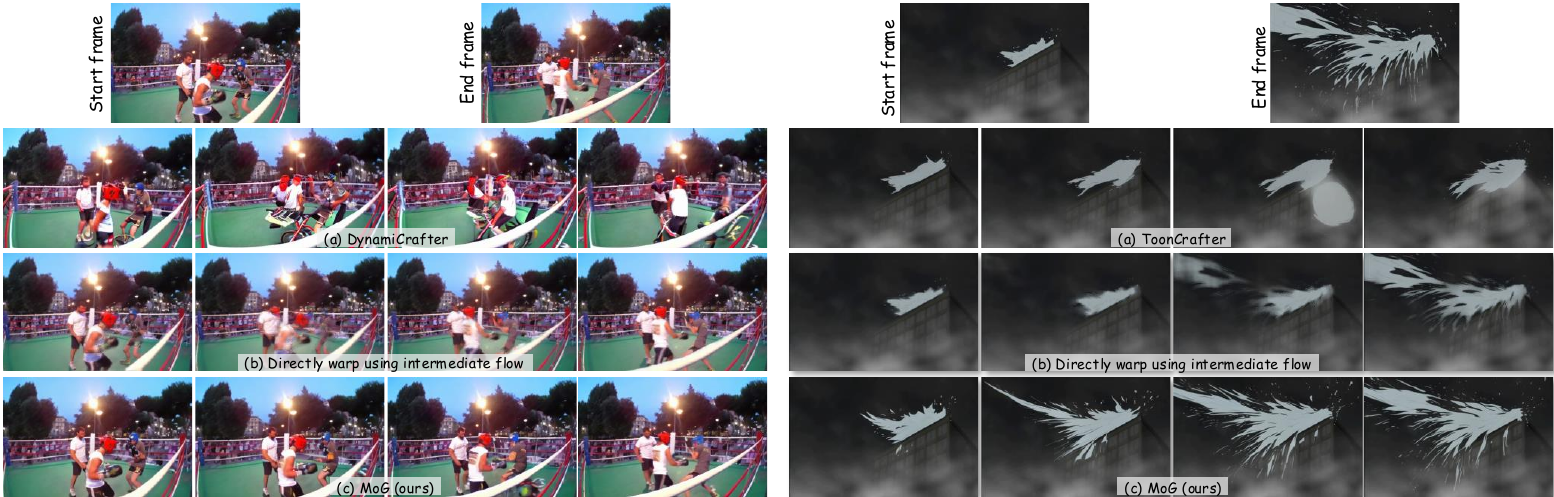}
  \vspace{-5mm}
  \caption{\textbf{Demonstration of MoG's guidance correction capability.} In both examples, DynamiCrafter or ToonCrafter struggle to generate temporally consistent motion in complex scenarios. While intermediate flow can provide valuable motion cues, it often introduces artifacts and fails to render fine appearance details. Leveraging the encoder-only injection design and selective parameter fine-tuning, MoG effectively integrates reliable motion information from intermediate flow while correcting its inaccuracies.}
  \label{fig:dynamic_correction}
  \vspace{-2mm}
\end{figure*}

\subsection{Task Definition}
For the input frames $x^1$ and $x^N \in \mathbb{R}^{3 \times H \times W}$, frame interpolation aims to generate a video comprising $N$ frames, denoted as $\mathbf{x} \in \mathbb{R}^{N \times 3 \times H \times W}$, where the first and last frames correspond to the input frames. 
\subsection{Intermediate Flow from Flow-based VFI} 
Flow-based methods explicitly estimate the correspondence between the starting and ending frames with respect to the intermediate frame, termed the intermediate flow. The intermediate flow can be obtained either by scaling the optical flow between  frames~\cite{hu2022many,niklaus2020softmax} or through direct prediction~\cite{RIFE,EMA-VFI}. In this work, we adopt the prediction-based method EMA-VFI~\cite{EMA-VFI}, owing to its versatility across various time steps and its task-oriented training~\cite{vimeo90k}.

Specifically, given the input frames $ x^1, x^N \in \mathbb{R}^{3 \times H \times W} $ as well as the $n$-th frame $x^n$ to be predicted, the intermediate flow $ f $ is computed using a learnable network \( \mathcal{O} \):
\begin{equation}
\label{eq:flow}
f_{1 \to n}, f_{N \to n}, M^n = \mathcal{O}(x^1, x^N,n).
\end{equation}
Here, $ f_{i \to n} \in \mathbb{R}^{2 \times H \times W} $ denotes the intermediate flow from the $i$-th frame $x^i$ to the $n$-th frame $x^n$, and $ M \in \mathbb{R}^{1 \times H \times W} $ represents the occlusion mask between the two frames at the $n$-th frame, taking values in the range of $ (0, 1) $. Subsequently, we can coarsely estimate the intermediate frame $ \bar{x}^n $ as follows:
\begin{equation}
\label{eq:warp}
\bar{x}^n = \text{warp}(x^1, f_{1 \to n}) \odot M^n + \text{warp}(x^N, f_{N \to n}) \odot (1 - M^n),
\end{equation}
where $ \text{warp}(x^i, f_{i \to n}) $ denotes the backward warping by $f_{i \to n}$, and $ \odot $ signifies the element-wise multiplication.

\subsection{VFI with Diffusion Models} 
 Empowered by large-scale pre-training, video diffusion models~\cite{SVD,dynamicrafter} exhibit remarkable capabilities in generating videos within complex scenarios. Recent works~\cite{GI,tooncrafter,dynamicrafter,TRF} have begun to leverage pre-trained video diffusion models for frame interpolation tasks. In this work, we explore our method based on two generative frame interpolation models, DynamiCrafter~\cite{dynamicrafter} and ToonCrafter~\cite{tooncrafter}, for real-world and animation scenes respectively.  Both models are based on Latent Diffusion Models (LDMs)~\cite{rombach2022high}, which conduct diffusion in the latent space of an autoencoder. Specifically, for any video \( \mathbf{x} \in \mathbb{R}^{N \times 3 \times H \times W} \), where \( N \) denotes the number of frames, the video is transformed into the latent space using a pre-trained encoder \( \mathcal{E}
 \) (\ie, VQ-VAE)~\cite{rombach2022high} to obtain the corresponding latent code $\mathbf{z}_0 = \mathcal{E}(\mathbf{x}) \in \mathbb{R}^{N \times C \times h \times w}$.  
 
 During training, $\mathbf{z}_0$ is first converted into an intermediate noisy video at timestep $ t $ using the equation:
\begin{equation}
\mathbf{z}_t = \alpha_t \mathbf{z}_0 + \sqrt{1 - \alpha_t} \mathbf{\epsilon}, \quad \mathbf{\epsilon} \sim \mathcal{N}(0, I).
\end{equation}
To achieve frame interpolation task,  a learnable denoising network \( \epsilon_\theta \) is then employed to predict the noise $\epsilon$ given the condition information from the first and the last frames. DynamiCrafter and ToonCrafter incorporate such condition information by:
\begin{equation}
\mathbf{\tilde{z}}_t = \left[\mathbf{z_t}; \mathbf{\bar{z}}_0\right], \quad \mathbf{\tilde{z}}_t \in \mathbb{R}^{N \times (2 \times C) \times h \times w},
\end{equation}
where $\mathbf{\bar{z}}_0$ is composed of the latent codes of bound frames $z_0^1,z_0^N$, while other positions remain zero. Then the denoising network is optimized by minimizing the following loss:
\begin{equation}
\label{eq:loss}
\mathcal{L}(\theta) = \mathbb{E}_{\mathbf{z}_0,t,\epsilon \sim \mathcal{N}(0,I)} \left[\left\|\epsilon-\epsilon_\theta\left(\mathbf{\tilde{z}}_t; t, c\right) \right\|_2^2\right].
\end{equation}
Here, \( c \) includes other condition information like the text and the fps. After training, we can iteratively recover \( \mathbf{\hat{z}}_0 \) using the input conditions and pure noise \( \mathbf{z}_T \sim \mathcal{N}(0, I) \), generating the video \( \mathbf{\hat{x}} = \mathcal{D}(\mathbf{\hat{z}}_0) \) via the decoder \( \mathcal{D} \).

The design of the denoising network \( \epsilon_\theta \) follows an U-Net-like structure~\cite{unet}, consisting of encoder blocks and decoder blocks. Each block comprises spatial and temporal layers. The spatial layers mainly consist of ResNet blocks~\cite{he2016res} and Transformer blocks~\cite{vaswani2017attention} with spatial attention, modeling spatial information within each frame, while the temporal layers are formed by Transformer blocks with temporal self-attention. 

\section{Motion-Aware Generative VFI}
\subsection{Motivation} 
\label{sec:motivation}
Flow-based frame interpolation methods can generate temporally smooth videos with the assistance of intermediate flow. 
However, in regions with complex motions, accurate intermediate flow estimation remains challenging due to the lack of sufficient supervisory signals—the annotation of real-world optical flow is prohibitively expensive. 
This limitation leads to severe artifacts such as blurring and ghosting in flow-based approaches.
In contrast, recent generative frame interpolation methods, benefiting from large-scale pre-training, demonstrate robust generative capabilities in complex scenarios. 
Unfortunately, they often suffer from incoherent motion and content inconsistencies with input frames. 
We attribute this to the fact that existing methods rely solely on generative models to infer motion trajectories between input frames—a capability insufficiently nurtured during generative pre-training.
To bridge the gap between flow-based stability and generative flexibility, we introduce Motion-aware Generative frame interpolation (MoG), as illustrated in \cref{fig:pipeline}. MoG employs the intermediate flow from flow-based methods as an explicitly motion guidance for generation, smoothing the motion of generated videos. Concurrently, MoG leverages the refinement capacities of generative models to automatically correct flow errors in complex motion regions.

\subsection{Dual Guidance Injection}
Since generative models cannot directly utilize the intermediate flow, it is imperative to devise a strategy for sufficiently injecting motion guidance into the denoising network. To this end, we propose dual guidance injection. Similar to the operation in \cref{eq:warp}, we coarsely estimate the representation of the intermediate frames using the intermediate flow. Subsequently, the estimated representations are seamlessly integrated into the model at both the latent and feature levels.

\paragraph{Latent-Level Injection.} To incorporate flow constraints at the latent level, we propose to inject additional latent codes of intermediate frames into the input. We backward-warp the latent codes of the input frames, guided by the intermediate flow. Specifically, given the latent codes of the start and end frames $ z_0^1 $ and $ z_0^N $, along with the motion guidance obtained through \cref{eq:flow}, we estimate the latent code of the $n$-th intermediate frame as follows:
\begin{equation}
\bar{z}_0^n = \text{warp}(z_0^1, f_{1 \to n}) \odot M^n + \text{warp}(z_0^N, f_{N \to n}) \odot (1 - M^n).
\end{equation}
Here, $ \bar{z}_0^n$ represents the estimated latent code for the $n$-th frame using the motion guidance. During training, the input to the denoising network is modified to:
\begin{equation}
\tilde{\mathbf{z}}_t = \left[\mathbf{z}_t; \bar{\mathbf{z}}_0\right], \quad \tilde{\mathbf{z}}_t \in \mathbb{R}^{N \times (2 \times C) \times h \times w}.
\end{equation}
It is noteworthy that no additional parameters are required, as our base models, DynamiCrafter or ToonCrafter, have already designed to accommodate extra inputs; In contrast, in their methods, the latent codes of intermediate frames in $ \mathbf{\bar{z}}_0 $ are always set to zero.

\paragraph{Feature-Level Injection.} To effectively integrate the motion guidance cross different granularities, we propose to inject guidance also in feature-level. Analogous to the latent-level, we estimate the features of intermediate frames based on the features $F^1, F^N \in \mathbb{R}^{D \times H \times W}$ of the input frames:
\begin{equation}
\bar{F}^i = \text{warp}(F^1, f_{1 \to i}) \odot M^n + \text{warp}(F^N, f_{N \to i}) \odot (1-M^n).
\end{equation}
In this equation, $\bar{F}^i$ represents the estimated features of the $i$-th frame guided by the intermediate flow. Unfortunately, unlike in latent-level injection, direct concatenation of the warped intermediate features into the network is not feasible. To address this issue, we reuse the temporal layer of the denoising network, which enables us to smooth and align the estimated intermediate features $\mathbf{\bar{F}}$ with the original feature distribution without introducing additional parameters. As shown in \cref{fig:pipeline}, the smoothed features $\mathbf{\hat{F}}$ is acquired by:
\begin{equation}
\mathbf{\hat{F}} = \text{Temporal layer}(\mathbf{\bar{F}}).
\end{equation}
Subsequently, we incorporate the flow-derived features $\dot{F}$ into the original features:
\begin{equation}
\mathbf{\tilde{F}} = \phi (\mathbf{F},\mathbf{\hat{F}} ).
\end{equation}
Remarkably, our exploration (refer to \cref{sec:ab}) reveals that a simple averaging already allows the generative model to effectively utilize the introduced motion guidance.

\subsection{Guidance Correction}
\label{sec:gc}
As mentioned in \cref{sec:motivation}, the intermediate flow cannot be accurately predicted in regions with complex motions. 
To alleviate the potential adverse effects, we have implemented two crucial designs to endow the generative model with the ability to automatically rectify error-prone regions.

Firstly, we adopt the design of encoder-only guidance injection. As depicted in \cref{fig:pipeline}, we introduce feature-level guidance injection exclusively in the encoder blocks of the denoising network. 
In this way, the decoder can appropriately adjust and rectify the information from the encoder, thereby capitalizing more effectively on the valuable information embedded in the flow guidance.
Secondly, we conduct selective parameter fine-tuning. We only fine-tune the spatial layers of all blocks. Given the inherent inaccuracy of the intermediate flow during the training phase, the model can learn how to dynamically harness the useful information within the guidance while concurrently acquiring the ability to repair flawed regions. Freezing the temporal layers serves to preserve the motion inference capability during pre-training and prevent the performance degradation associated with the feature-level guidance injection.

\paragraph{Discussion.} To validate the effectiveness of our proposed method, as shown in \cref{fig:dynamic_correction}, we showcase two examples for a comparison among the results derived from the intermediate flow, those of the original generative model, and the results produced by MoG.
Evidently, MoG can fully exploit the motion cues encapsulated within the intermediate flow to generate videos with better temporal smoothness. Meanwhile, in regions with complex motions, MoG rectifies the inaccuracies from the intermediate flow and leverages the generative capabilities to yield more reasonable appearance details. More quantitative comparisons can be found in \cref{sec:ab}.

\section{Experiment}

\begin{table*}[h]
\centering
\renewcommand{\arraystretch}{1.02}
\resizebox{1.0\textwidth}{!}{
\begin{tabular}{lcccccccccccccc}
\Xhline{1.0pt}
\multirow{2.4}{*}{\textbf{Models}} & \multicolumn{2}{c}{\textbf{PSNR}~($\uparrow$)} & \multicolumn{2}{c}{\textbf{SSIM}~($\uparrow$)} & \multicolumn{2}{c}{\textbf{LPIPS}~($\downarrow$)}  & \multicolumn{2}{c}{\textbf{FID}~($\downarrow$)} & \multicolumn{2}{c}{\textbf{CLIP$_{\text{sim}}$}~($\uparrow$)} & \multicolumn{2}{c}{\textbf{FVD}~($\downarrow$)} & \multicolumn{2}{c}{\textbf{VBench}~($\uparrow$)}\\
\cmidrule(lr){2-3} \cmidrule(lr){4-5} \cmidrule(lr){6-7} \cmidrule(lr){8-9} \cmidrule(lr){10-11} \cmidrule(lr){12-13} \cmidrule(lr){14-15} & Real & Anime & Real & Anime & Real & Anime & Real & Anime & Real & Anime & Real & Anime & Real & Anime \\
\Xhline{0.8pt}
\multicolumn{13}{l}{\textbf{\textit{Flow-based VFI models}}} \\
RIFE~\cite{RIFE} & 18.21 & 20.33 & 0.5672 & 0.7587 & 0.3601 & 0.3407 & 55.27 & 62.35 & 0.8304 & 0.8693 & 742.29 & 628.55 & 77.57 & 78.59\\
EMA-VFI~\cite{EMA-VFI} & 18.17 & 20.49 & 0.5731& 0.7531 & 0.3619 & 0.3701 & 51.09 & 53.27 & 0.8411 & 0.8907 & 717.58 & 517.60 & 78.15 & 80.02\\
\Xhline{0.8pt}
\multicolumn{13}{l}{\textbf{\textit{Generative VFI models}}} \\
LDMVFI~\cite{ldmvfi} & 17.17 & 18.39 & \textbf{0.5953} & 0.7175 & 0.3081 & 0.2860 & 41.47 & 46.18 & 0.8703 & 0.8710 & 479.63 & 435.17  & 78.91 & 80.21 \\
GI~\cite{GI} & 15.95 & 18.04 & 0.5271 & 0.6971 & 0.3384 & 0.2891 & 36.06 & 46.18 & 0.8703 & 0.8710 & 521.00 & 449.31  & 79.97 & 81.97 \\
TRF~\cite{TRF} & 15.43 & 16.49 & 0.5132 & 0.6744 & 0.3920 & 0.3470 & 42.48 & 53.95 & 0.8491 & 0.8731 & 624.63 & 481.02  & 79.01 & 80.79\\
DynamiCrafter~\cite{dynamicrafter}  & 16.05 & -- & 0.5225 & -- & 0.3380 & --  & 42.16 & -- & 0.8634 & -- & 562.34 & -- & 79.51 & -- \\
ToonCrafter~\cite{tooncrafter}  & -- & 18.01 & -- & 0.7182 & -- & 0.2944 & -- & 40.63 & -- & 0.9203 & -- & 425.71  & -- & 82.57\\
\rowcolor{blue!10} \textbf{MoG (ours)} & \textbf{17.82} & \textbf{19.44} & 0.5898 & \textbf{0.7434} & \textbf{0.2716} & \textbf{0.2615} & \textbf{31.26} & \textbf{33.73} & \textbf{0.9083} & \textbf{0.9320} & \textbf{401.49} & \textbf{351.41} & \textbf{81.44} & \textbf{83.31}\\
\Xhline{1.0pt}
\end{tabular}
}
\vspace{-1mm}
\caption{\textbf{Quantitative comparison on VFIBench.} }
\vspace{-2mm}
\label{tab:gt_metric_cmp}

\end{table*}

\subsection{Implementation Details}
We develop MoG based on DynamiCrafter~\cite{dynamicrafter} for real-world scenes and ToonCrafter~\cite{tooncrafter} for animation scenes. MoG employs EMA-VFI~\cite{EMA-VFI} for intermediate flow prediction. For model fine-tuning, we only train the spatial layers, while keeping all other parameters fixed. We train with the same loss in \cref{eq:loss} for 20K steps on $1\times10^{-5}$ learning rate and batch size 32.  The training dataset is internal collected of $512\times320$ resolution with 16 frames. The sampling strategy is consistent with~\cite{dynamicrafter} and~\cite{tooncrafter}.

\subsection{VFIBench}
To evaluate interpolated frames, we present VFIBench, a comprehensive benchmark that encompasses diverse data, including real-world videos and animations. It employs various metrics for a detailed assessment of frame quality and fidelity to ground truth. VFIBench also poses a challenge by requiring models to interpolate 14 frames between specified start and end frames. This setup demands advanced motion modeling capabilities. For data collection, we meticulously selected 100 samples from the DAVIS 2017 dataset~\cite{pont20172017}, referred to as the VFIBench-Real, to reflect real-world scenarios. Additionally, we curate another set of 100 samples from internet animations, called VFIBench-Ani, which includes a diverse range of styles from Japanese, American, and Chinese animations.

A well-interpolated video should not only be of high quality inherently but also maintain fidelity to the ground truth. For the former, we adopt six metrics from VBench~\cite{huang2024vbench}: subject consistency, background consistency, temporal flickering, motion smoothness, aesthetic quality, and imaging quality. The average performance across all metrics is reported as VBench in  \cref{tab:gt_metric_cmp}. These metrics collectively assess the intrinsic quality. For the latter, we employ six widely adopted metrics: PSNR, SSIM, LPIPS~\cite{zhang2018unreasonable}, FID~\cite{heusel2017gans}, and the CLIP similarity score~\cite{CLIP} for image-level comparison, and FVD~\cite{unterthiner2018towards, unterthiner2019fvd} for video-level comparison. 

\begin{figure*}[t]
  \centering
  \includegraphics[width=0.91\textwidth]{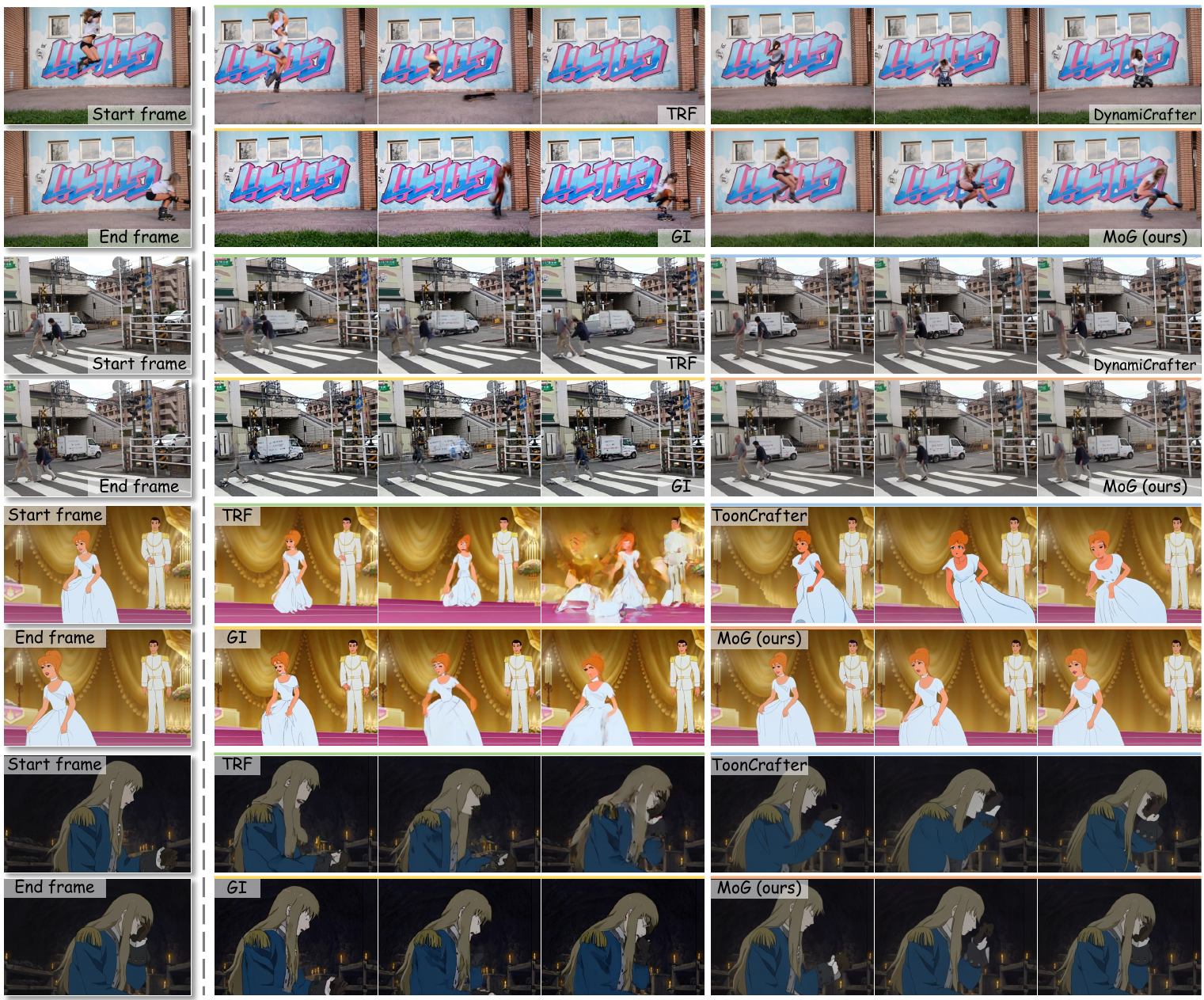}
  \vspace{-2mm}
  \caption{\textbf{Visual comparison on real-world and animation scenes.}}
  \label{fig:qualitative_cmp_main}
  \vspace{-2pt}
\end{figure*}

\subsection{Main Results and Analysis}
We evaluate our MoG by benchmarking it against state-of-the-art methods across two categories: flow-based interpolation methods, specifically RIFE~\cite{RIFE} and EMA-VFI~\cite{EMA-VFI}, and generative interpolation methods, including LDMVFI~\cite{ldmvfi}, GI~\cite{GI}, TRF~\cite{TRF}, DynamiCrafter~\cite{dynamicrafter} and ToonCrafter~\cite{tooncrafter}. \textit{To conduct a more equitable comparison, we retrain the flow-based methods and LDMVFI~\cite{ldmvfi} on our dataset.} Note that DynamiCrafter and ToonCrafter are tailored for real-world and  animation, respectively. Hence their performance is reported separately for each domain.

\paragraph{Quantitative results.}
As shown in \cref{tab:gt_metric_cmp}, compared to others generative VFI methods, MoG exhibits significant improvements in video quality and fidelity to ground truth. This indicates that our method can effectively utilize the intermediate flow to generate smooth motion, and meanwhile, it has successfully reduced the erroneous information within it. Compared to flow-based VFI models, our approach also demonstrates notable enhancements across most metrics; however, it lags in PSNR. We argue that this discrepancy arises because flow-based VFI often produces blurry results in complex motion scenarios (as illustrated in \cref{fig:teaser}), which can inflate these metrics while compromising actual visual quality~\cite{zhang2018unreasonable}. We also conduct a comparison on the publicly available dataset Davis-7~\cite{vidim}. As shown in \cref{tab:davis}, MoG achieves best performance across all metrics.

\begin{table}[t]
\centering
\resizebox{0.8\columnwidth}{!}{
    \begin{tabular}{lccc}
    \Xhline{1.0pt}
\textbf{Davis-7} &  \textbf{SSIM} ($\uparrow$) & \textbf{LPIPS} ($\downarrow$) & \textbf{FID} ($\downarrow$) \\
 \Xhline{0.8pt}
LDMVFI~\cite{ldmvfi} &  0.4175	& 0.2765 & 56.28\\
VIDIM~\cite{vidim} &  0.4221 & 0.2986 & 53.38\\
DynamiCrafter~\cite{dynamicrafter} &  0.4785 & 0.3752 & 75.06\\
\rowcolor{blue!10} \textbf{MoG (ours)} &  \textbf{0.5978} & \textbf{0.2641} & \textbf{51.94}\\
\Xhline{1.0pt}
    \end{tabular}
}
\vspace{-2mm}
 \caption{\textbf{Comparsion on Davis-7~\cite{vidim}.}}
 \label{tab:davis}
\end{table}

\begin{table}[t]
\centering
\vspace{-2mm}
\renewcommand{\arraystretch}{1.05}
\resizebox{0.45\textwidth}{!}{
\begin{tabular}{lcccc}
\Xhline{1.0pt}
Methods & \textbf{\Centerstack{Motion\\Quality}} & \textbf{\Centerstack{Temporal\\Coherence}} & \textbf{\Centerstack{Frame\\Fidelity}} & \textbf{\Centerstack{Overall\\Quality}}\\
\Xhline{0.8pt}
EMA-VFI~\cite{EMA-VFI} & 0.49\% & 0.74\% & 0.49\% & 0.49\%\\
TRF~\cite{TRF} & 1.73\% & 1.73\% & 0.99\% & 1.48\% \\
GI~\cite{GI} & 16.05\% & 14.57\% & 23.21\% & 15.56\% \\
DynamiCrafter~\cite{dynamicrafter} & 3.70\% & 3.21\% & 4.20\% & 2.22\%\\
\rowcolor{blue!10} \textbf{MoG (ours)} & \textbf{78.02}\% & \textbf{79.75}\% & \textbf{71.11}\% & \textbf{80.25}\%\\
\Xhline{1.0pt}
\end{tabular}
}
\vspace{-2mm}
\caption{\textbf{User study statistics.}}
\label{tab:user_study}
\vspace{-2mm}
\end{table}

\begin{table}[t]
    \centering
    \setlength\tabcolsep{2pt}
    \resizebox{1.0\columnwidth}{!}{
    \begin{tabular}{lcccc}
    \Xhline{1.0pt}
 & GI~\cite{GI} & TRF~\cite{TRF} & DynamiCrafter~\cite{dynamicrafter} & MoG (ours) \\
 \Xhline{0.8pt}
\textbf{Runtime (s)} & 385.19 & 141.41 & \textbf{33.42} & 34.08 \\ 
\textbf{\textbf{VBench}~($\uparrow$)} & 79.97 & 79.01 & 79.51 & \textbf{81.44} \\ 
\Xhline{1.0pt}
    \end{tabular}
    }\vspace{-2mm}
    \caption{\textbf{Comparsion on computational efficiency.}}
\vspace{-4mm}
\label{tab:compute}
\end{table}

\paragraph{Qualitative results.}
We present qualitative comparisons with three generative VFI methods in \cref{fig:qualitative_cmp_main}. Lacking explicit motion guidance, these methods struggle to accurately infer and understanding the correspondences between input frames, resulting in inconsistent content and unstable motion. In contrast,  MoG achieves superior motion and visual quality in complex scenarios. More comparisons are available in supplementary materials.
\paragraph{User study.}
 To further verify the advantage of our method, we also conduct a comprehensive user study. Participants are instructed to select the best-generated videos based on motion quality, temporal coherence, frame fidelity, and overall quality. We collect results from 27 participants and report the findings in \cref{tab:user_study}. Thanks to the explicit motion guidance, the study shows a clear preference for our method in all aspects.
\paragraph{Computational efficiency .}
We further compare the computational efficiency of different generative frame interpolation methods. As shown in \cref{tab:compute}, the testing resolution is unified as $16\times 512\times 320$. Compared with TRF and GI, MoG not only exhibits superior performance but also significantly improves computational efficiency. When compared with our baseline model, DynamicCrafter, MoG have substantially enhanced the quality of the generated videos while only slightly increasing the computational load.

\subsection{Ablation Study}
For brevity, we only conduct ablation experiments in real-world scenarios. Our analysis primarily relies on four metrics to evaluate different strategies: two pertaining to video quality, namely Subject Consistency and Background Consistency (abbreviated as \textbf{Sub. Cons.} and \textbf{Bg. Cons.} in \cref{tab:ab_all}), and two metrics assessing fidelity between the video and ground truth, specifically LPIPS and FVD.
\label{sec:ab}

\begin{table*}[t]
\label{tab:ab_all}
\centering
% \hfill
\begin{subtable}[t]{0.45\linewidth}
\centering
\renewcommand{\arraystretch}{1.1}
\footnotesize
\begin{tabular}{c|cccc}
 & \textbf{Sub. Cons.} & \textbf{Bg. Cons.} & \textbf{LPIPS} & \textbf{FVD} \\
\Xhline{0.8pt}
Only fine-tuning & 89.57 & 92.09 & 0.3290 & 540.47 \\
Linear interpolation & 90.77 & 92.75 & 0.3046 & 481.41\\
Pretrained optical flow & 91.52 & 93.44 & 0.2871 & 454.29 \\
\rowcolor{blue!10} \textbf{Flow-based VFI} & \textbf{92.65} & \textbf{94.34} & \textbf{0.2716} & \textbf{401.49} \\
\end{tabular}
\vspace{-1mm}
\caption{\textbf{Choice of motion guidance.}}
\label{tab:select_which_guidance}
\end{subtable}
\vspace{1mm}
\hfill
\begin{subtable}[t]{0.45\linewidth}
\centering
\renewcommand{\arraystretch}{1.1}
\footnotesize
\begin{tabular}{cc|cccc}
Latent & Feature & \textbf{Sub. Cons.} & \textbf{Bg. Cons.} & \textbf{LPIPS} & \textbf{FVD} \\
\Xhline{0.8pt}
  & & 89.57 & 92.09 & 0.3290 & 540.47\\
\ding{51} & & 91.87 & 93.92 & 0.2796 & 437.85\\
 & \ding{51} & 92.34 & 93.74 & 0.2811 & 424.50 \\
\rowcolor{blue!10} \ding{51} & \ding{51} & \textbf{92.65} & \textbf{94.34} & \textbf{0.2716} & \textbf{401.49} \\
\end{tabular}
\vspace{-1mm}
\caption{\textbf{Different levels of guidance injection.}}
\label{tab:ablation_injection_points}
\end{subtable}
\vspace{1mm}
\hfill
\\
% \hfill
\begin{subtable}[t]{0.45\linewidth}
\centering
\renewcommand{\arraystretch}{1.1}
\footnotesize
\begin{tabular}{c|cccc}
 & \textbf{Sub. Cons.} & \textbf{Bg. Cons.} & \textbf{LPIPS} & \textbf{FVD}\\
 \Xhline{0.8pt}
 Transformers & 92.35 & 94.01 & 0.2750 & 431.21 \\
 Convolution & 92.44 & 94.09 & 0.2745 & 426.85 \\
 Linear & 92.49 & 94.17 & 0.2739 & 422.97 \\
 \rowcolor{blue!10} \textbf{Average} & \textbf{92.65} & \textbf{94.34} & \textbf{0.2716} & \textbf{401.49} \\
\end{tabular}
\caption{\textbf{Different ways to merge guidance.}}
\label{tab:analysis_merge_strategy}
\end{subtable}
\hfill
\begin{subtable}[t]{0.45\linewidth}
\centering
\renewcommand{\arraystretch}{1.1}
\footnotesize
\begin{tabular}{c|cccc}
 & \textbf{Sub. Cons.} & \textbf{Bg. Cons.} & \textbf{LPIPS} & \textbf{FVD} \\
 \Xhline{0.8pt}
 All & 92.17 & 93.51 & 0.2792 & 451.31\\
Decoder-only & 91.74 & 92.97 & 0.2942 & 471.52\\
\rowcolor{blue!10} \textbf{Encoder-only} & \textbf{92.65} & \textbf{94.34} & \textbf{0.2716} & \textbf{401.49}\\
\end{tabular} 
\caption{\textbf{Position of feature-level injection.}}
\label{tab:ablation_latent_feature_inject}
\end{subtable}
\caption{\textbf{Ablation experiments.} The \colorbox[HTML]{E6E6FD}{colored background} indicates our default setting.}
\vspace{-5mm}
\end{table*}

\begin{figure}[t]
  \centering
  \includegraphics[width=0.49\textwidth]{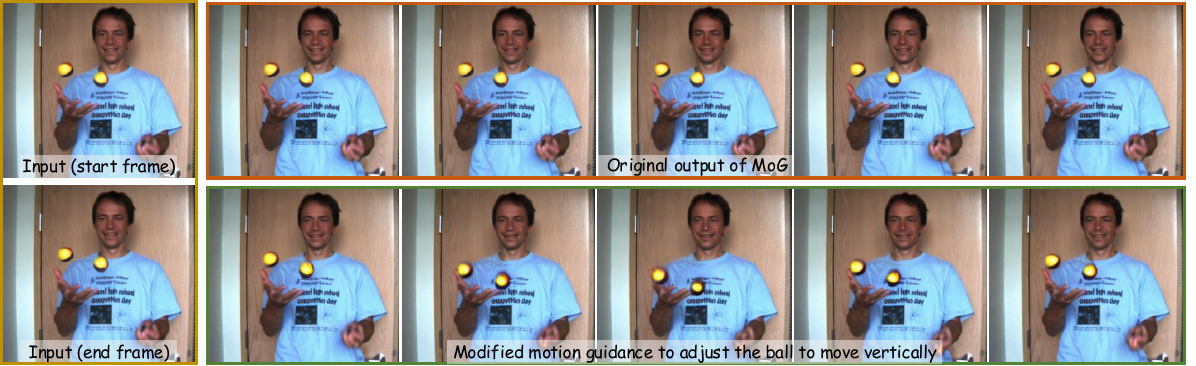}
  \vspace{-5mm}
  \caption{\textbf{An example of manually altering intermediate flow.}}
  \label{fig:toy_exmaple}
  \vspace{-5mm}
\end{figure}

\paragraph{Effectiveness of intermediate flows.} To validate that intermediate flows can indeed impose motion constraints on generative models, we attempt to manually modify the intermediate flows to control the motion. As depicted in  \cref{fig:toy_exmaple}, given the same start and end frames, MoG would generate a nearly static video by default. When we manually alter the intermediate flows of the ball to  move vertically, the generated frames change accordingly. 

\begin{figure}[t]
  \centering
  \includegraphics[width=0.43\textwidth]{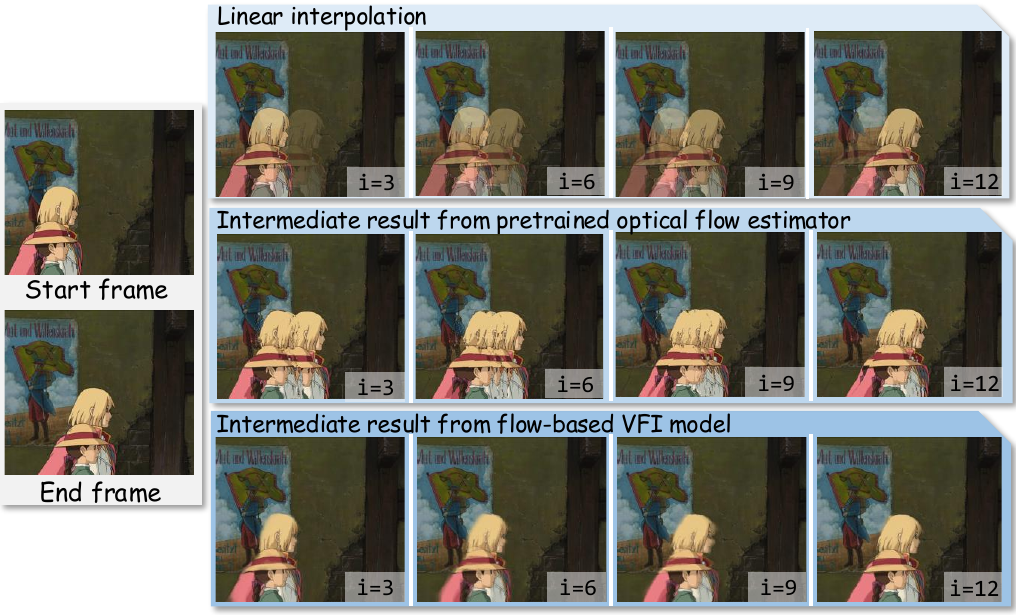}
  \vspace{-2mm}
  \caption{\textbf{Visualizations on different methods as guidance.} }
  \label{fig:flow_comparsion}
  \vspace{-3mm}
\end{figure}

In addition, we explore two other potential forms of motion guidance. One is the linear interpolation of the input frames, and the other is the flow from a pre-trained optical flow estimator~\cite{xu2022gmflow}. As shown in  \cref{fig:flow_comparsion}, the intermediate flows exhibit smoother motion and fewer errors. We attribute this to the fact that intermediate flows are task-oriented flows~\cite{vimeo90k}, which are more suitable for the frame interpolation. Moreover, the predicted occlusion masks in  \cref{eq:flow} can also enhance the accuracy of the guidance. The performance advantages presented in  \cref{tab:select_which_guidance} further verify this conclusion.

\paragraph{Dual guidance injection.} We introduce motion guidance at both the latent and feature levels. To validate the effectiveness of each level, we compare the performance with motion guidance introduced at only one level, or without motion guidance, as in \cref{tab:ablation_latent_feature_inject}. The results demonstrate that the injection of either level all significantly enhances video quality and fidelity metrics. Specifically, latent-level injection is more beneficial for background consistency, while feature-level injection improves subject consistency. The best performance is achieved when both levels are employed, allowing the generative model to leverage motion guidance at different granularities.

Regarding the design of feature-level injection, we also attempt methods that first concatenate the warped intermediate features with the original features, followed by learnable modules such as Transformer blocks~\cite{vaswani2017attention}, convolutions, or linear layers. Surprisingly, as shown in \cref{tab:analysis_merge_strategy}, a simple averaging yielded the best performance, possibly due to the substantial data requirements for the learnable modules to achieve strong generalization.

\paragraph{Guidance correction.} We conduct comprehensive ablation experiments on designs proposed in \cref{sec:gc}. First, regarding the integration location of motion guidance, we evaluate three configurations: guidance injection into all blocks (All), exclusive injection into decoder blocks (Decoder-only), and sole injection into encoder blocks (Encoder-only). As demonstrated in \cref{tab:ablation_injection_points},  Encoder-only achieves the optimal performance, while Decoder-only yields the lowest performance. This discrepancy aligns with our hypothesis that maintaining decoder guidance-free allows the model to effectively rectify flow-constrained information from the encoder, ensuring the generated videos adhere more closely to pre-trained distributions.

\begin{figure}[t]
  \centering
  \includegraphics[width=0.47\textwidth]{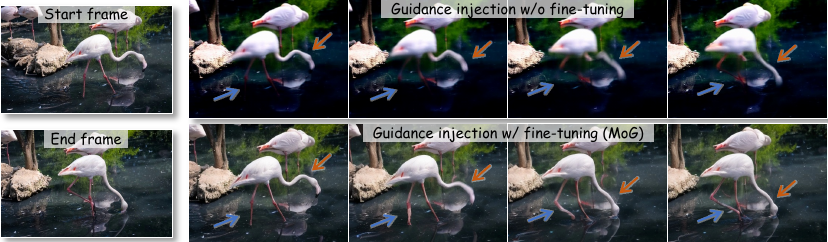}
  \vspace{-2mm}
  \caption{\textbf{Visualizations on selective parameter fine-tuning.} }
  \label{fig:finetune}
  \vspace{-4mm}
\end{figure}

Furthermore, as illustrated in \cref{fig:finetune}, we validate the effectiveness of selective parameter fine-tuning. Results show that while intermediate flow guidance can improve temporal smoothness, it still leads to artifacts in complex regions. Selective fine-tuning substantially enhances the generated video quality by enabling dynamic guidance adaptation in spatial layers, while preserving the motion generation capabilities through frozen temporal layers. 
% In terms of where to merge motion guidance, we experiment with three configurations: injecting into all blocks (All), exclusively into decoder blocks (Decoder-only), and solely into encoder blocks (Encoder-only). As illustrated in \cref{tab:ablation_injection_points}, the Encoder-only achieves the best performance, while Decoder-only results in the poorest performance. This discrepancy may stem from the fact that modifications to the decoder can disrupt the powerful video generation capabilities from pre-training.

\section{Conclusion} In this work, we present a novel generative frame interpolation framework, MoG, which simultaneously enhances the motion smoothness by intermediate flows and adaptively correcting flow errors through generative refinement. We first propose dual guidance injection to introduce flow constraints into the generative models at both the latent and feature levels. Then we conduct encoder-only guidance injection and selective parameters fine-tuning for guidance correction. Through extensive experiments in both real-world and animated scenes, we demonstrate that MoG achieves substantial improvements in video quality and fidelity. 
\section*{Appendix}

\subsection*{A. More Visual Comparisons} 
\begin{figure*}[t]
  \centering
  \includegraphics[width=0.8\textwidth]{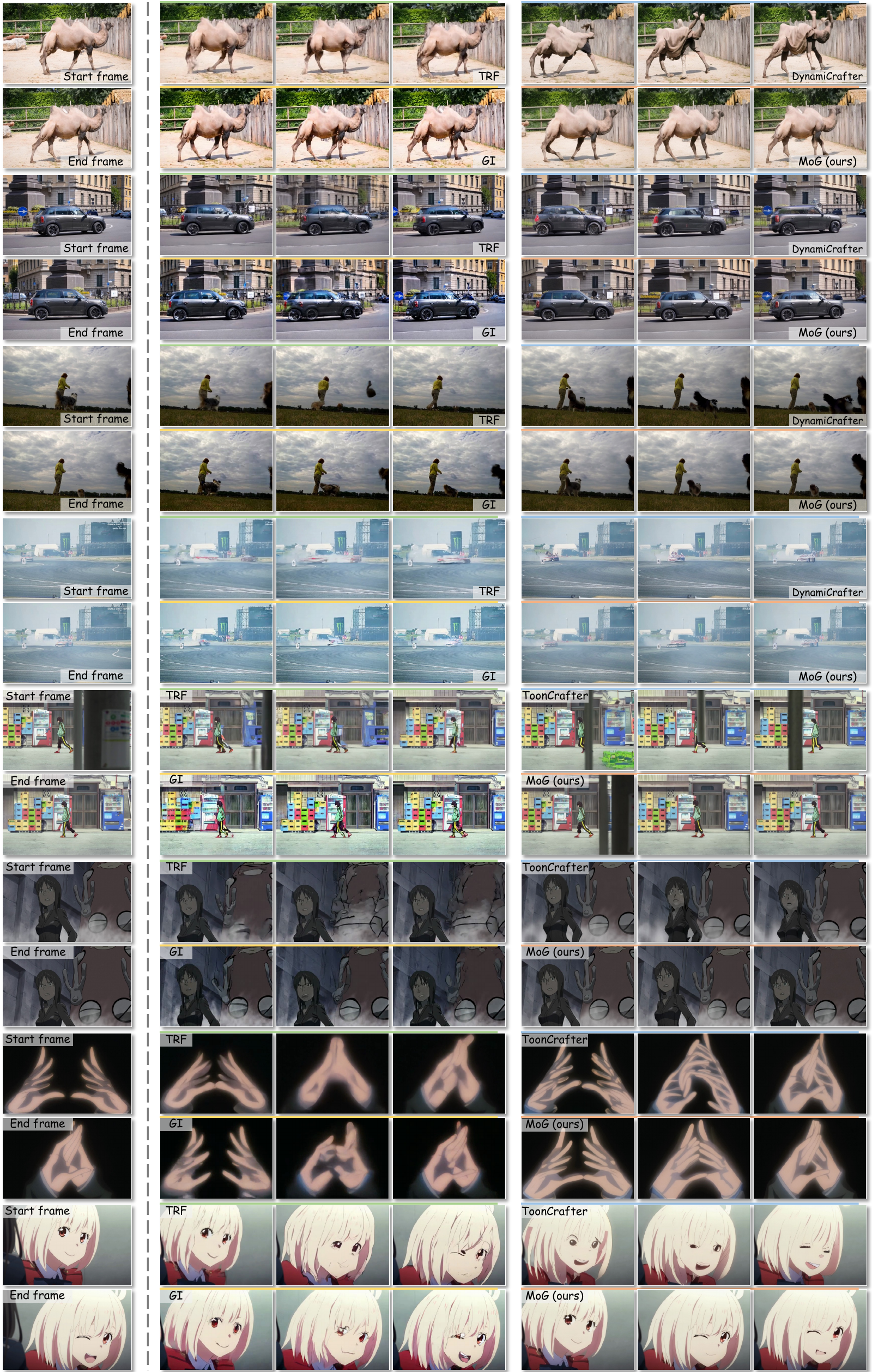}
  \vspace{-3mm}
  \caption{\textbf{Additional qualitative comparison on real-world and animation scenes.}}
  \label{fig:supp_qualitative_cmp}
\end{figure*}

To further demonstrate the improvement of our method, we also provide additional qualitative comparisons in~\cref{fig:supp_qualitative_cmp}.

\subsection*{B. Limitations and Future Work}
Despite MoG has achieved non-trivial improvement in generation quality across various scenes, there still several limitations warrant further exploration. Firstly, our approach is built upon the U-Net architecture of the DynamiCrafter model.  However, the video generation capabilities of DynamiCrafter has lagged behind recently DiT-based~\cite{peebles2023scalable} video generation models~\cite{yang2024cogvideox,mochi1}, which constrains our performance ceiling. Unfortunately, we currently lack the necessary resources and data to work with the latest models. Investigating our method within the new models presents a promising direction for future work. Secondly, MoG relies on the flow-based VFI model, meaning that the quality of its outputs may impact the effectiveness of motion guidance. Enhancing the generalizability of flow-based VFI across diverse scenes will also benefit our method moving forward.

% \newpage
{
    \small
    \bibliographystyle{ieeenat_fullname}
    \bibliography{main}
}

\end{document}